\documentclass[runningheads]{llncs}

\usepackage[T1]{fontenc}
\usepackage{graphicx}
\usepackage{amsfonts}
\usepackage{amssymb}
\usepackage{multirow}
\usepackage{tabularx}
\usepackage{graphicx}
\usepackage{marvosym}
\usepackage{booktabs}
\usepackage{array}
\usepackage[colorlinks, linkcolor=blue, citecolor=blue, urlcolor=blue]{hyperref}
\usepackage{color}

\begin{document}

\title{
MULTI-SEMANTIC FUSION MODEL FOR GENERALIZED ZERO-SHOT SKELETON-BASED ACTION RECOGNITION\thanks{Supported by the National Key R\&D Program of China (2022ZD0117901), the National Natural Science Foundation of China (62106260, 62236010, 62076078, U19B2036 and 62225601), the Beijing Natural Science Foundation Project (Z200002), the Program for Youth Innovative Research Team of BUPT (2023QNTD02), and the High-performance Computing Platform of BUPT.}
}
\titlerunning{MULTI-SEMANTIC FUSION MODEL FOR GZSSAR}
\author{Ming-Zhe Li\inst{1,3}, Zhen Jia\inst{3}, Zhang Zhang\inst{2,3}\textsuperscript{(\Letter)}, Zhanyu Ma\inst{1}, \\ and Liang Wang\inst{2,3}}
\authorrunning{MZ. Li et al.}
\institute{
PRIS Lab., School of Artificial Intelligence, \\ Beijing University of Posts and Telecommunications, Beijing, China \\
\email{\{limingzhe\_24,mazhanyu\}@bupt.edu.cn} \and
School of Artificial Intelligence, University of Chinese Academy of Sciences (UCAS, Beijing, China
\and
CRIPAC, MAIS, CASIA, Beijing, China \\
\email{\{zhen.jia,zzhang,wangliang\}@nlpr.ia.ac.cn}
}

\maketitle

\vspace{-0.3cm}
\begin{abstract}
Generalized zero-shot skeleton-based action recognition (GZ-SSAR) is a new challenging problem in computer vision community, which requires models to recognize actions without any training samples. Previous studies only utilize the action labels of verb phrases as the semantic prototypes for learning the mapping from skeleton based actions to a shared semantic space. However, the limited semantic information of action labels restricts the generalization ability of skeleton features for recognizing unseen actions. In order to solve this dilemma, we propose a multi-semantic fusion (MSF) model for improving the performance of GZSSAR, where two kinds of class-level textual descriptions ($i.e.$, action descriptions and motion descriptions), are collected as auxiliary semantic information to enhance the learning efficacy of generalizable skeleton features. Specially, a pre-trained language encoder takes the action descriptions, motion descriptions and original class labels as inputs to obtain rich semantic features for each action class, while a skeleton encoder is implemented to extract skeleton features. Then, a variational autoencoder (VAE) based generative module is performed to learn a cross-modal alignment between skeleton and semantic features. Finally, a classification module is built to recognize the action categories of input samples, where a seen-unseen classification gate is adopted to predict whether the sample comes from seen action classes or not in GZSSAR. The superior performance compared with previous models validates the effectiveness of the proposed MSF model on GZSSAR. (The code has been released at \href{https://github.com/EHZ9NIWI7/MSF-GZSSAR}{MSF-GZSSAR})


\end{abstract}

\vspace{-0.3cm}
\section{Introduction}
\vspace{-0.3cm}
Human action recognition is a fundamental computer vision problem with great application potentials on video surveillance~\cite{aggarwal2011human}, video retrieval~\cite{poppe2010survey} and human-computer interaction~\cite{weinland2011survey}.
With the developments of advanced depth cameras and pose estimation agorithms, skeleton-based action recognition has become a hot research topic, besides the ordinary RGB video based action recognition.
Skeleton-based action recognition takes the 3D skeleton sequences as input, which makes action recognition models more robust to deal with variations in illumination, camera viewpoints and other background changes.
However, although there have been high-performance skeleton-based approaches~\cite{cheng2020skeleton,liu2020disentangling,shi2019two,zhang2019view}, most of these approaches are prone to overfitting and fail to generalize to the unseen classes outside the training set~\cite{gupta2021syntactically}. 
This is mainly because the conventional action recognition only attempts to learn a mapping that maximizes the inter-class distance, and may not be suitable of learning generalizable features for encoding new action categories.

\vspace{-0.7cm}
\begin{figure}[ht]
\centering
\includegraphics[width=\textwidth]{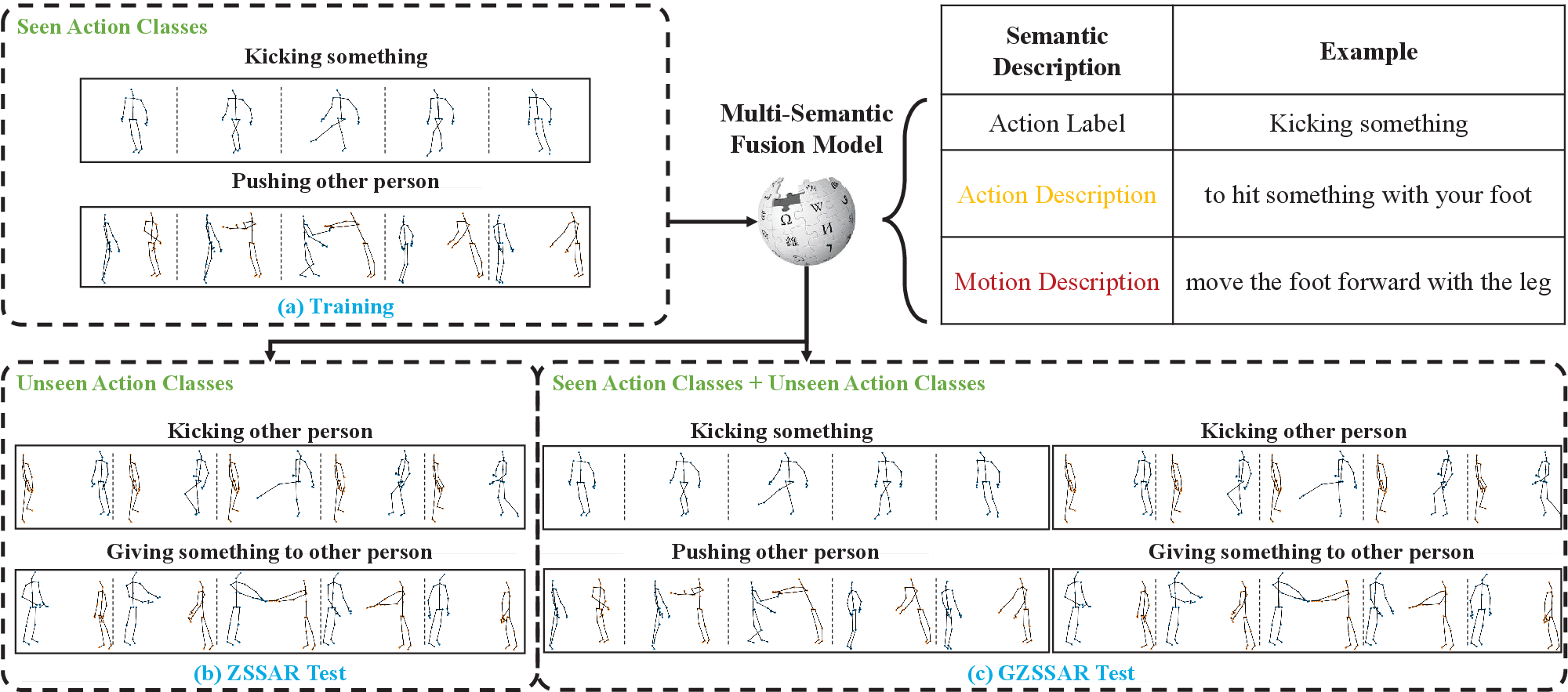}
\vspace{-7mm}
\caption{A schematic diagram of ZSSAR and GZSSAR.} \label{fig1}
\end{figure}
\vspace{-0.7cm}

Therefore, there is a strong motivation for exploring the problems of zero-shot skeleton-based action recognition (ZSSAR) and generalized zero-shot skeleton-based action recognition (GZSSAR) methods.
In the previous work~\cite{gupta2021syntactically}, Gupta et al. set up the task of ZSSAR and GZSSAR, as the schematic diagram shown in Fig.~\ref{fig1}.
ZSSAR aims to train a model that can classify actions of unseen classes via transferring knowledge obtained from other seen classes with the help of semantic information. 
The setting of GZSSAR needs model to recognize samples from both seen and unseen classes simultaneously during test time, which is a more challenging issue and closer to open-world applications. 

Generally, in generalized zero-shot image classification~\cite{pourpanah2022review}, the semantic information, such as language and attributes, is utilized to build a relationship between seen and unseen classes through mapping visual features and semantic features into a common space. 
However, the previous methods~\cite{liu2021goal,zhang2020towards} are difficult to be directly adopted to GZSSAR.
On one hand, there are not any semantic attribute based annotations or detailed language descriptions for action categories in current skeleton-based action datasets.
On the other hand, as the only semantic information that can be directly utilized, the class labels which commonly are verb phrases, are too ambiguous to well distinguish the actions in skeleton modality. 
For example, action ``reading book'' and action ``typing keyboard'' are very similar in motion.
To deal with the dilemma, auxiliary semantic descriptions are collected to better describe actions in words.
The semantic descriptions include action descriptions and motion descriptions.
The action descriptions are obtained by referring to the explanations of actions in the Oxford Dictionary. 
The motion descriptions are collected by human annotators who are asked to describe the movements of body parts when watching some examples of each action category.
The class-level annotations can be completed easily with low manual costs. 
In this paper, the two kinds of semantic descriptions as well as the class labels are investigated to construct rich semantic representations so as to improve the learning efficacy of generalizable skeleton features.
To the best of our knowledge, we are the first to explore rich semantic descriptions for GZSSAR.

\vspace{-0.7cm}
\begin{figure}[ht]
\centering
\includegraphics[width=\textwidth]{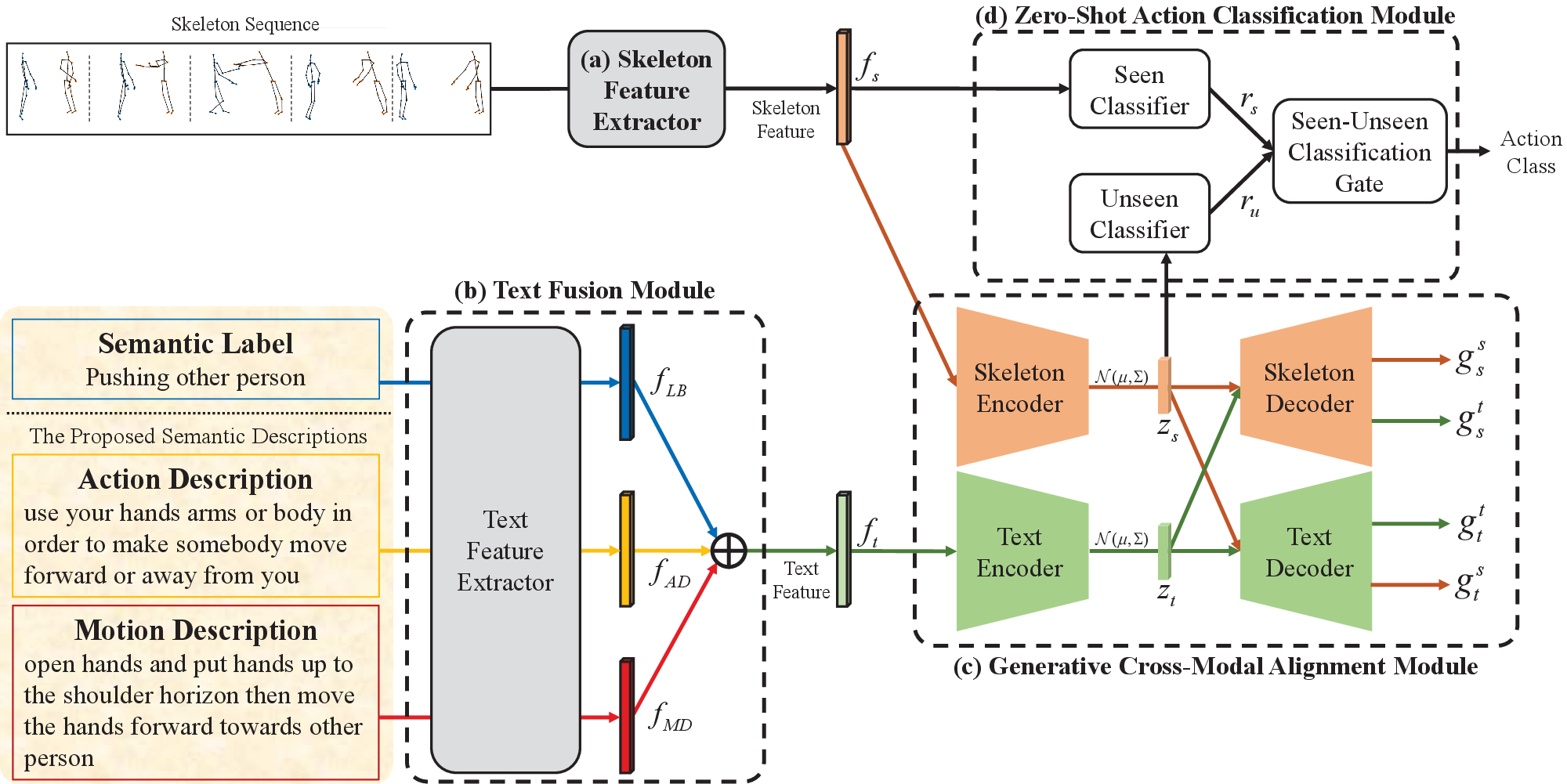}
\vspace{-7mm}
\caption{Architectural diagram for the proposed Multi-Semantic Fusion model.} \label{fig2}
\end{figure}
\vspace{-0.7cm}

Since action descriptions are from textual knowledge in dictionary and motion descriptions describe the detailed movements of body parts,
we design the multi-semantic fusion (MSF) model to comprehensively utilize multiple semantic information to improve the performance of GZSSAR.
The overall architecture of the MSF model is shown in Fig.\ref{fig2}.
Firstly, for the input skeleton sequences, the MSF model uses a 4s-ShiftGCN~\cite{cheng2020skeleton} as the skeleton feature extractor to encode skeleton features.
Secondly, for the input multiple semantic information, a text fusion module is designed to encode each type of semantic information and fuse them to obtain rich semantic features for each action categories.
A pre-trained ViT-B/32~\cite{radford2021learning} is implemented as the text feature extractor.
Thirdly, considering the superiority of the generative methods in generalized zero-shot image classification, a VAE-based generative cross-modal alignment module is designed to map the skeleton features and the rich semantic features to a common feature space. 
Therefore, the model is able to generate latent embeddings of unseen classes for the training of an unseen-classifier.
At last, in order to achieve better GZSSAR performance, a zero-shot action classification module is implemented at the end of the model. 
It uses a seen-unseen classification gate to make a decision that the given sample should be classified as seen classes or unseen classes in the task setting of GZSSAR. 
The results of the experiments prove the effectiveness of the MSF model and the superiority of the proposed semantic descriptions.

The contributions of this paper can be summarized in threefold:

$\bullet$ Two kinds of class-level descriptions ($i.e.$, action descriptions and motion descriptions) are firstly explored for GZSSAR. 
The richer semantic information is validated effective for improving the performance of GZSSAR.  

$\bullet$ A multi-semantic fusion model is proposed to comprehensively utilize multiple semantic information and superiority accomplish GZSSAR task. A VAE-based generative module is adopted to accomplish the cross-modal alignment between skeleton features and rich semantic features of actions. 

$\bullet$ Extensive experiments are performed to demonstrate the proposed MSF model's state-of-the-art performance on NTU-60~\cite{shahroudy2016ntu} and NTU-120~\cite{liu2019ntu} datasets.

\vspace{-0.4cm}
\section{Related Work}
\vspace{-0.2cm}

\label{related work}

\subsubsection{Generalized Zero-Shot Skeleton-Based Action Recognition} 
In recent years, there are only a few studies related to GZSSAR due to its immense difficulties in bridging the semantic gap between skeleton data and semantic labels without any prior exposure to training samples of unseen classes. 
Jasani et al.~\cite{jasani2019skeleton} propose to use embedding based methods~\cite{frome2013devise,sung2018learning} to align skeleton embeddings with their corresponding text embeddings of the class labels ($e.g.$, ``take off jacket'' and ``put on glasses'').
Gupta et al.~\cite{gupta2021syntactically} propose a model termed SynSE, which splits a class label into verbs and nouns according to its syntactic structures ($e.g.$, ``drink water'' can be splitted into a verb ``drink'' and a noun ``water'') and aligns the skeleton embeddings with the corresponding verb embeddings and noun embeddings.
They also formally set up the task of zero-shot learning (ZSL) and generalized zero-shot learning (GZSL) in skeleton-based action recognition.

\vspace{-0.5cm}
\subsubsection{Semantic Information in Generalized Zero-Shot Learning}
The most widely used semantic information for GZSL can be grouped into manually defined attributes~\cite{lampert2013attribute} and word vectors~\cite{mikolov2013distributed}. 
Manually defined attributes describe the high-level characteristics of classes ($e.g.$, shape and color), which enable the models to recognize classes that never appear in the training dataset. 
The attribute space has been widely used in GZSL studies, however it requires human efforts in attribute annotations for each classes, which are not suitable for large-scale problems~\cite{kodirov2017semantic}. 
Along with the developments of foundation language models, word vector based methods can directly map the class name into a semantic embedding space and describe the similarities between seen classes and unseen ones in a more efficient way. 
Therefore, they are more suitable for large-scale datasets.
However, for zero-shot action recognition, the class names may not reflect the detailed differences between various actions. 
For example, the actions ``reading book'' and ``typing keyboard'' are both related to the movement of two hands. 
The simple class names are not semantically abundant enough to describe the ambiguity in the two classes.
Thus, in this paper, we attempt to augment the descriptions of action categories with richer semantic information.

\vspace{-0.3cm}
\section{Action and Motion Descriptions for GZSSAR}
\vspace{-0.3cm}
\label{semantic information}
In generalized zero-shot image classification, the word vectors of class labels or visual attributes are used as semantic information. 
However, accurately representing action classes in a semantic space is still a challenging problem due to the complexity and diversity of human actions.
Furthermore, for skeleton-based action recognition, there is not any other semantic information in current popular skeleton-based action datasets ($e.g.$, NTU-60 and NTU-120 datasets), except for class labels.
However, the class labels ($e.g.$, ``drink water'' and ``eat meal'') are always too concise and abstract to well distinguish the actions in most cases.
As shown in the previous work, the performance of GZSSAR is still limited due to the lack of transferable semantic information among different actions.
To address this dilemma, action descriptions and motion descriptions are explored for GZSSAR in this paper. The examples of action descriptions and motion descriptions are demonstrated in Table~\ref{text examples}.

\vspace{-0.5cm}
\begin{table}[ht]
\scriptsize
\centering
\caption{Action and Motion Description Examples}
\label{text examples}
\vspace{-2.5mm}
\begin{tabularx}{\textwidth}{>{\centering\arraybackslash}m{0.3\textwidth}>{\centering\arraybackslash}m{0.33\textwidth}>{\centering\arraybackslash}m{0.33\textwidth}}
\toprule
\textbf{\textbf{Class Labels}} & \textbf{\textbf{Action Descriptions}} & \textbf{\textbf{Motion Descriptions}} \\ \midrule
brushing teeth & to clean polish or make teeth smooth with a brush & move the hand up to the head then tremble the hand \\ \cmidrule(lr){1-3}
pickup & to take hold of something and lift it up & bend knees and hips to move the hand down to the ground then pinch then get up \\ \cmidrule(lr){1-3}
wear on glasses & to have glasses on your head as a piece of decoration & pinch and move hands up to the head then release hands and put hands down \\ \cmidrule(lr){1-3}
salute & to touch the side of your head with the fingers of your right hand to show respect & open the hand and move the hand to the side of the head then hold \\ \cmidrule(lr){1-3}
drop & to allow something to fall by accident from your hand & release the hand in front of the middle of the spine \\ \bottomrule
\end{tabularx}
\end{table}
\vspace{-0.7cm}

As shown in Table~\ref{text examples}, the action descriptions are a kind of textual descriptions about actions, which are collected with reference to the explanations of actions in the Oxford Dictionary~\cite{stevenson2010oxford}. 
For all action categories in the NTU-60 and NTU-120 datasets, we search for the definitions of verb phrases in class labels, and form the action descritions for all action categories.
Compared to class labels, the action descriptions are easier to understand and obviously contain human knowledge on various action categories which is crucial for GZSSAR.
However, an unavoidable issue is that the definitions of an action category phrases may include other new abstract semantic concepts which are hard to be grounded to input skeleton samples.
Therefore, the redundant new concepts may not bring benefits for the performance of GZSSAR.

Thus, we further propose the motion descriptions which are annotated with reference to the movement trajectories of relatively important body parts in skeleton sequences.
To collect the annotations, we invite 10 human annotators who are graduate students in an engineering institute to describe the movement processes of all action categories in NTU-60 and NTU-120 datasets. 
The annotators are asked to use some simple and common verbs ($e.g.$, ``move'' and ``put'') as well as the nouns of 25 key-points ($e.g.$, ``head'', ``shoulder'' and ``elbow'') to describe the movements of important key-points as watching a skeleton sequence sample randomly selected from each action category.
In the motion descriptions, a set of simple verbs and nouns which directly correspond to certain body parts can be shared across various action categories. 
Compared with class labels and action descriptions, the shared descriptions on basic movements of body parts may be more appropriate for learning transferable skeleton features across seen and unseen classes.

In summary, the proposed action descriptions contain richer vocabulary from human knowledge base.
Meanwhile, the motion descriptions can better reflect the detailed movements of important skeleton joints in actions.
The proposed action descriptions and motion descriptions will be available to the GZSSAR research community in the future. 

\vspace{-0.3cm}
\section{Methodology}
\vspace{-0.2cm}
\subsection{Problem Formulation}
\vspace{-0.1cm}
Assume that $S = \left\{(x_{i}^{s},t_{i}^{s},y_{i}^{s})_{i=1}^{N_{s}}|x_{i}^{s}\in{X^{s}}, t_{i}^{s}\in{T^{s},y_{i}^{s}\in{Y^{s}}} \right\}$ represents the seen action class dataset, and $U = \left\{(x_{j}^{u},t_{j}^{u},y_{j}^{u})_{j=1}^{N_{u}}|x_{j}^{u}\in{X^{u}}, t_{j}^{u}\in{T^{u},y_{j}^{u}\in{Y^{u}}} \right\}$ denotes the unseen action class dataset.
$x_{i}^{s}, x_{j}^{u}\in{\mathbb{R}^{D}}$ indicate the $D$-dimensional skeleton features in skeleton feature space $X$, and $t_{i}^{s}, t_{j}^{u}\in{\mathbb{R}^{K}}$ indicate the $K$-dimensional text features in text feature space $T$. 
$N_{s}$ and $N_{u}$ are the numbers of seen and unseen samples.
$X^{u} = \{ x_j^u \}_{j=1}^{N_u}$ indicate the skeleton feature set of unseen action classes in $X$.
$Y^{s} = \left\{y_{1}^{s},...,y_{C_{s}}^{s} \right\}$ and $Y^{u} = \left\{y_{1}^{u},...,y_{C_{u}}^{u} \right\}$ indicate the digital label sets of seen and unseen action classes in $Y$, where $C_{s}$ and $C_{u}$ are the numbers of seen and unseen action classes. 
$Y={Y^{s}}\cup{Y^{u}}$ denotes all action classes and ${Y^{s}}\cap{Y^{u}}={\varnothing}$. 
The objective of ZSSAR and GZSSAR are to learn $f_{ZSSAR}: {X^{u}}\to{Y^{u}}$ and $f_{GZSSAR}:{X}\to{Y}$, respectively.

\vspace{-0.4cm}
\subsection{Skeleton Feature Extractor}
\vspace{-0.2cm}
\label{skeleton extractor}
To complete the GZSSAR task, a crucial requirement is to acquire the discriminative features of skeleton sequences.
Considering the superior performance of ShiftGCN~\cite{cheng2020skeleton} in skeleton based action recognition, the proposed MSF model adopts ShiftGCN as the skeleton feature extractor. 
The skeleton feature extractor in Fig.\ref{fig2} (a) and the seen-classifier in Fig.\ref{fig2} (d) need to be trained on the train-set ($S_{train}$) first in order to recognize samples in the test-set ($S_{test}$).
$S_{train}$ and $S_{test}$ are two complementary and disjoint subsets partitioned from $S$.

\vspace{-0.4cm}
\subsection{Text Fusion Module}
\vspace{-0.2cm}
As introduced in Section~\ref{semantic information}, 
the action descriptions and the motion descriptions are complementary to each other.
To comprehensively utilize these rich semantic information, the text fusion module takes multiple semantic information ($i.e.$, class labels, action descriptions and motion descriptions) as input and extracts their semantic features separately. 
Then, a concatenation operation is adopted to fuse the three types of semantic information together, as shown in Fig.\ref{fig2} (b). 
Since CLIP~\cite{radford2021learning} is effective in image-text matching tasks, the MSF model directly adopts the pre-trained ViT-B/32, which is the text-encoder of CLIP, as the text feature extractor to extract semantic features.
The MSF model uses a simple and efficient way ($i.e.$, concatenation operation) to fuse multiple semantic information together, as shown in Equation~\ref{concat}. 

\vspace{-0.4cm}
\begin{equation}
\label{concat}
    f_t = f_{LB} \oplus f_{AD} \oplus f_{MD}.
\end{equation}
\vspace{-0.6cm}

\noindent{$\oplus$ indicates the concatenation operation. $f_{LB}$, $f_{AD}$ and $f_{MD}$ indicate the features of the class label, action description and motion description respectively.}

\vspace{-0.45cm}
\subsection{Generative Cross-Modal Alignment Module}
\vspace{-0.2cm}
\label{generative module}
Since the skeleton feature $f_{s}$ and the text feature $f_{t}$ belong to two different feature spaces, it is necessary to map the $f_{s}$ and the $f_{t}$ to a common feature space where they are aligned with each other.
For this, we employ a generative variational autoencoder (VAE) based architecture~\cite{gupta2021syntactically} to implement the mapping. The architecture of the generative cross-modal alignment module is shown in Fig.\ref{fig2} (c). 
The module contains two branches, $i.e.$, the skeleton branch and the text branch.
Since the structure of the module is symmetrical, only the skeleton branch is introduced in detail here. 
In the training phase, the evidence lower bound loss (ELBO) is implemented for the training of the VAE. Formally,

\vspace{-0.3cm}
\begin{equation}
    \mathcal{L}_{VAE}^{s} = \mathbb{E}_{q_{\phi}(z_{s}|f_{s})}[\log{p_{\theta}(f_{s}|z_{s})}]-{\beta}D_{KL}(q_{\phi}(z_{s}|f_{s})\parallel p_{\theta}(z_{s}|f_{s})),
\end{equation}
\vspace{-0.5cm}

\noindent{where $p_{\theta}(\cdot)$ and $q_{\phi}(\cdot)$ denote the likelihood and the prior respectively. $\beta$ is a hyper-parameter which acts as a trade-off factor between the two error terms. $q_{\phi}(z_{s}|f_{s})$ obeys the multivariate Gaussian distribution $\mathcal{N}_{s}(\mu_{s}, \Sigma_{s})$.
The L2 loss is implemented to align the generated $g_{t}^{s}$ with the text feature $f_t$. Formally,}

\vspace{-0.3cm}
\begin{equation}
\mathcal{L}_{Align}^{s} =  \parallel{f_{t}-g_{t}^{s}}\parallel_{2},
\end{equation}
\vspace{-0.5cm}

\noindent{where $\parallel{\cdot}\parallel_{2}$ denotes the L2 norm. The entire loss function of the skeleton branch is formulated as follows.}

\vspace{-0.5cm}
\begin{equation}
    \mathcal{L}^{s} = \mathcal{L}_{VAE}^{s} + {\alpha}\mathcal{L}_{Align}^{s},
\end{equation}
\vspace{-0.65cm}

\noindent{where $\alpha$ is a trade-off weight factor. 

Similarly, the entire loss function of the text branch $\mathcal{L}^{t} = \mathcal{L}_{VAE}^{t} + \mathcal{L}_{Align}^{t}$ is calculated in text branch.
Finally, the entire loss function of the generative cross-modal alignment module is formulated as follows.}

\vspace{-0.4cm}
\begin{equation}
    \mathcal{L} = \mathcal{L}^{s} + \mathcal{L}^{t}.
\end{equation}
\vspace{-0.65cm}

Through this way, the encoders and the decoders are trained relying on $\mathcal{L}_{VAE}^{s}$ and $\mathcal{L}_{VAE}^{t}$. The alignment between $z_s$ and $z_t$ is achieved with the help of $\mathcal{L}_{align}^{s}$ and $\mathcal{L}_{align}^{t}$. 
After training, the generative cross-modal alignment module is able to generate latent embeddings from the input feature, in an aligned feature space.

\vspace{-0.45cm}
\subsection{Zero-Shot Action Classification Module}
\vspace{-0.15cm}
As shown in Fig.~\ref{fig2} (d), the zero-shot action classification module is composed of the seen-classifier, the unseen-classifier, and the seen-unseen classification gate. 
As mentioned in Section~\ref{skeleton extractor}, the seen-classifier has already been trained with the skeleton feature extractor. In ZSSAR, an unseen-classifier is still needed to classify the samples in $U$. Since samples in $X^{u}$ can not be used for training, we generate $z_{t}$ using $t_{j}^{u}\in{T^{u}}$ 
to train the unseen-classifier. 

In GZSSAR, a common method is to utilize an additional classifier to classify the test samples from both seen and unseen classes.
However, the prediction results obtained by this way are usually unsatisfying~\cite{atzmon2019adaptive}. 
Therefore, a seen-unseen classification gate is implemented to predict the action classes of the input skeleton features more accurately. 
The classification gate is trained on a validation set, utilizing the predicted results from the seen-classifier and the unseen-classifier as inputs. It employs a logistic regression classifier to regress binary results, indicating whether the input skeleton features belong to unseen action classes or not. The validation set is partitioned from the training set based on the number of unseen classes.

\vspace{-0.35cm}
\section{Experimental Results}
\vspace{-0.15cm}
\subsection{Datasets}
\vspace{-0.1cm}

\subsubsection{NTU RGB+D 60~\cite{shahroudy2016ntu}}
The NTU-60 dataset is a large-scale indoor dataset for 3D human action analysis. It contains 56,880 human action videos collected by three Kinect-V2 cameras. The dataset consists of 60 action classes. 
Only the skeleton data is used in this work. 
In each skeleton sequence, every frame contains no more than 2 skeletons, and each skeleton is composed of 25 joints.
Two seen/unseen splits ($i.e.$, 55 (seen classes)/5 (unseen classes) and 48 (seen classes)/12 (unseen classes)) are set up in the previous work~\cite{gupta2021syntactically}, in where the unseen classes are chosen randomly.
In order to compare with the state-of-the-art methods, we continue to adopt the two seen/unseen splits. 
The selection of the unseen classes maintains the same with the previous work~\cite{gupta2021syntactically}.

\vspace{-0.5cm}
\subsubsection{NTU RGB+D 120~\cite{liu2019ntu}}
The NTU-120 dataset is currently the largest indoor skeleton-based action recognition dataset, which is an extended version of the NTU-60 dataset. It contains a total of 114,480 videos performed by 106 subjects from 155 viewpoints. The dataset consists of 120 classes, extended from the 60 classes of the NTU-60 dataset. 
The seen/unseen splits are 110 (seen)/10 (unseen) and 96 (seen)/24 (unseen), also the same with the previous work~\cite{gupta2021syntactically}.

\vspace{-0.35cm}
\subsection{Implementation Details}
\vspace{-0.15cm}
The training phase can be divided into 4 stages, training of the skeleton feature extractor, training of the generative cross-modal alignment module, training of the unseen-classifier and training of the seen-unseen classification gate.
In stage 1, the skeleton feature extractor and the seen-classifier are trained using samples of seen classes to obtain 256-dimensional skeleton features. 
The specific training details of the skeleton feature extractor are the same as~\cite{cheng2020skeleton}. 
The 512-dimensional text features are obtained through the pre-trained text feature extractor. 
In stage 2, the generative cross-modal alignment module is trained using the skeleton features and the text features. 
The dimension of the latent embeddings is set to 100 on the NTU-60 dataset and 200 on the NTU-120 dataset. 
The optimizer is Adam with the learning rate 1$\times e^{-4}$, the training epoch is 1900 with the batch size of 64. In stage 3, we generate 500 latent features for each unseen class to train the unseen-classifier for 300 epochs. 
The learning rate is set to 1$\times e^{-3}$ with Adam optimizer. 
In stage 4, the logistic regression classifier is optimized using LBFGS solver with the default aggressiveness hyper-parameter (C = 1). ~\cite{atzmon2019adaptive}. 
All our experiments are performed on one NVIDIA GTX TITAN X GPU.

\vspace{-0.35cm}
\subsection{Comparisons With State-of-the-art Methods}
\vspace{-0.15cm}
Since there have been a few previous work for ZSSAR and GZSSAR, Gupta et al.~\cite{gupta2021syntactically} modify representative generalized zero-shot image classification methods and implement them for GZSSAR from scratch. 
In this part, the proposed MSF model is compared with the state-of-the-art method SynSE~\cite{gupta2021syntactically} and the other modified GZSL methods. 
The evaluation protocol of the MSF model and the SynSE maintains the same.
The experiments of both ZSSAR and GZSSAR are conducted.
In ZSSAR experiments, the accuracy of the classification for unseen samples is reported.
And in GZSSAR, the accuracy of seen classes ($Acc_s$), the accuracy of unseen classes ($Acc_u$) and their harmonic mean ($H$) are all reported.

\vspace{-0.5cm}
\begin{table}[ht]
\centering
\scriptsize
\caption{Comparisons with SOTA methods in ZSSAR accuracies (\%).}
\label{SOTA ZSL}
\vspace{-2.5mm}
\begin{tabular}{@{}ccccc@{}}
\toprule
\multirow{2.6}{*}{\textbf{Method}} & \multicolumn{2}{c}{ \textbf{NTU-60}} & \multicolumn{2}{c}{ \textbf{NTU-120}} \\ \cmidrule(lr){2-3} \cmidrule(lr){4-5}
 & 55($s$)/5($u$) & 48($s$)/12($u$) & 110($s$)/10($u$) & 96($s$)/24($u$) \\ \midrule
ReViSE~\cite{hubert2017learning} & 53.91 & 17.49 & 55.04 & 32.38 \\
JPoSE~\cite{wray2019fine} & 64.32 & 28.75 & 51.93 & 32.44 \\
CADA-VAE~\cite{schonfeld2019generalized} & 76.84 & 28.96 & 59.53 & 35.77 \\
SynSE~\cite{gupta2021syntactically} & 75.81 & 33.30 & 62.69 & 38.70 \\ \cmidrule(lr){1-5}
Ours(LB) & 80.01 & 40.80 & 67.33 & 47.57 \\
Ours(AD) & 78.91 & 45.81 & 56.03 & 42.99 \\
Ours(MD) & 83.26 & \textbf{55.38} & 60.36 & 45.39 \\
Ours(LB+AD+MD) & \textbf{83.63} & 49.19 & \textbf{71.20} & \textbf{59.73} \\ \bottomrule
\end{tabular}
\end{table}

\vspace{-1.2cm}
\begin{table}[ht]
\centering
\scriptsize
\caption{Comparisons with SOTA methods in GZSSAR accuracies (\%) and harmonic mean.}
\label{SOTA GZSL}
\vspace{-2.5mm}
\begin{tabular}{@{}ccccccccccccc@{}}
\toprule
\multirow{4.3}{*}{\textbf{Method}} & \multicolumn{6}{c}{\textbf{NTU-60}} & \multicolumn{6}{c}{\textbf{NTU-120}} \\ \cmidrule(lr){2-7} \cmidrule(lr){8-13}
 & \multicolumn{3}{c}{55($s$)/5($u$)} & \multicolumn{3}{c}{48($s$)/12($u$)} & \multicolumn{3}{c}{110($s$)/10($u$)} & \multicolumn{3}{c}{96($s$)/24($u$)} \\ \cmidrule(lr){2-4} \cmidrule(lr){5-7} \cmidrule(lr){8-10} \cmidrule(lr){11-13}
 & $Acc_s$ & $Acc_u$ & $H$ & $Acc_s$ & $Acc_u$ & $H$ & $Acc_s$ & $Acc_u$ & $H$ & $Acc_s$ & $Acc_u$ & $H$ \\ \midrule
ReViSE~\cite{hubert2017learning} & 74.22 & 34.73 & 29.22 & 62.36 & 20.77 & 31.16 & 48.69 & 44.84 & 46.68 & 49.66 & 25.06 & 33.31 \\
JPoSE~\cite{wray2019fine} & 64.44 & 50.29 & 56.49 & 60.49 & 20.62 & 30.75 & 47.66 & 46.40 & 47.05 & 38.62 & 22.79 & 28.67 \\
CADA-VAE~\cite{schonfeld2019generalized} & 69.38 & 61.79 & 65.37 & 51.32 & 27.03 & 35.41 & 47.16 & 19.78 & 48.44 & 41.11 & 34.14 & 37.31 \\
SynSE~\cite{gupta2021syntactically} & 61.27 & 56.93 & 59.02 & 52.21 & 27.85 & 36.33 & 52.51 & 57.60 & 54.94 & 56.39 & 32.25 & 41.04 \\ \cmidrule(lr){1-13}
Ours(LB) & 69.41 & 57.15 & 62.69 & 53.25 & 34.43 & 41.82 & 56.45 & 58.38 & \textbf{57.40} & 58.96 & 35.71 & 44.48 \\
Ours(AD) & 67.34 & 60.69 & 63.84 & 59.42 & 37.52 & 46.00 & 49.87 & 52.87 & 51.33 & 59.66 & 33.45 & 42.87 \\
Ours(MD) & 65.04 & \textbf{66.74} & 65.88 & 50.69 & \textbf{48.75} & \textbf{49.70} & 58.67 & 52.38 & 55.35 & 58.76 & 32.86 & 42.15 \\
Ours(LB+AD+MD) & 71.73 & 66.15 & \textbf{68.83} & 58.80 & 40.00 & 47.61 & 46.84 & \textbf{68.30} & 55.57 & 56.84 & \textbf{48.61} & \textbf{52.40} \\ \bottomrule
\end{tabular}
\end{table}
\vspace{-0.5cm}

As shown in Table~\ref{SOTA ZSL} and Table~\ref{SOTA GZSL}, in both ZSSAR and GZSSAR, the proposed MSF model surpasses all the state-of-the-art methods by a large margin.
Even if only using class labels, the MSF model achieves better performance than SynSE, due to the use of CLIP based text feature extractor. 
Noted, the skeleton feature extractor used in this work, $i.e.$, ShiftGCN, is also the same with the SynSE.
Compared to SynSE, the most significant increase of MSF in ZSSAR accuracy reaches up to 21.03\%, and the most significant increase of MSF in $H$ reaches up to 11.36\%.
The increase of $Acc_u$ in GZSSAR is particularly significant.
Furthermore, in the 48($s$)/12($u$) split of NTU-60 and 96($s$)/24($u$) split of NTU-120, our method achieves greater performance improvements, which verifies the advantages of MSF on more challenging data splits.

\vspace{-0.35cm}
\subsection{Ablation Studies}
\vspace{-0.1cm}
\subsubsection{Comparisons of Text Feature Extractors:}
In CLIP, there are two pre-trained text encoders that are frequently used in previous vision-language models ($i.e.$, ViT-B/16 and ViT-B/32). 
Both of them can be directly used as the text feature extractor. 
To select one of them for feature extraction, both of the two encoders are tested in ZSSAR and GZSSAR.
The results are presented in Table~\ref{ext ZSL} and Table~\ref{ext GZSL}. 
As shown in the two tables, in both ZSSAR and GZSSAR, ViT-B/32 performs better than ViT-B/16 in most cases. 
Only under the 96($s$)/24($u$) split of NTU-120 dataset, the performance of ViT-B/32 is slightly inferior to ViT-B/16.
Therefore, ViT-B/32 is more suitable as the text feature extractor in the MSF model. 
In the subsequent experiments, we will use ViT-B/32 as the text feature extractor.

\vspace{-0.5cm}
\begin{table}[ht]
\centering
\scriptsize
\caption{Comparisons of different text feature extractors in ZSSAR accuracies (\%).}
\label{ext ZSL}
\vspace{-2.5mm}
\begin{tabular}{@{}ccccc@{}}
\toprule
\multirow{2.6}{*}{\textbf{Model}} & \multicolumn{2}{c}{\textbf{NTU-60}} & \multicolumn{2}{c}{\textbf{NTU-120}} \\ \cmidrule(lr){2-3} \cmidrule(lr){4-5}
 & 55($s$)/5($u$) & 48($s$)/12($u$) & 110($s$)/10($u$) & 96($s$)/24($u$) \\ \midrule
ViT-B/16 & 83.55 & 43.98 & 69.34 & \textbf{60.29} \\
ViT-B/32 & \textbf{83.63} & \textbf{49.19} & \textbf{71.20} & 59.73 \\ \bottomrule
\end{tabular}
\end{table}

\vspace{-1.2cm}
\begin{table}[ht]
\centering
\scriptsize
\caption{Comparisons of different text feature extractors in GZSSAR accuracies (\%) and harmonic mean.}
\label{ext GZSL}
\vspace{-2.5mm}
\begin{tabular}{@{}ccccccccccccc@{}}
\toprule
\multirow{4.3}{*}{\textbf{Model}} & \multicolumn{6}{c}{\textbf{NTU-60}} & \multicolumn{6}{c}{\textbf{NTU-120}} \\ \cmidrule(lr){2-7} \cmidrule(lr){8-13}
 & \multicolumn{3}{c}{55($s$)/5($u$)} & \multicolumn{3}{c}{48($s$)/12($u$)} & \multicolumn{3}{c}{110($s$)/10($u$)} & \multicolumn{3}{c}{96($s$)/24($u$)} \\ \cmidrule(lr){2-4} \cmidrule(lr){5-7} \cmidrule(lr){8-10} \cmidrule(lr){11-13}
 & $Acc_s$ & $Acc_u$ & $H$ & $Acc_s$ & $Acc_u$ & $H$ & $Acc_s$ & $Acc_u$ & $H$ & $Acc_s$ & $Acc_u$ & $H$ \\ \midrule
ViT-B/16 & 71.85 & 65.49 & 68.52 & 55.04 & 36.73 & 44.06 & 46.07 & 67.03 & 54.61 & 57.14 & \textbf{49.90} & \textbf{53.27} \\
ViT-B/32 & 71.73 & \textbf{66.15} & \textbf{68.83} & 58.80 & \textbf{40.00} & \textbf{47.61} & 46.84 & \textbf{68.30} & \textbf{55.57} & 56.84 & 48.61 & 52.40 \\ \bottomrule
\end{tabular}
\end{table}

\vspace{-1.2cm}
\begin{table}[ht]
\centering
\scriptsize
\caption{Comparisons of different semantic information in ZSSAR accuracies (\%).}
\label{sem ZSL}
\vspace{-2.5mm}
\begin{tabular}{@{}ccccc@{}}
\toprule
\multirow{2.6}{*}{\textbf{Method}} & \multicolumn{2}{c}{\textbf{NTU-60}} & \multicolumn{2}{c}{\textbf{NTU-120}} \\ \cmidrule(lr){2-3} \cmidrule(lr){4-5}
 & 55($s$)/5($u$) & 48($s$)/12($u$) & 110($s$)/10($u$) & 96($s$)/24($u$) \\ \midrule
LB & 80.01 & 40.80 & 67.33 & 47.57 \\
AD & 78.91 & 45.81 & 56.03 & 42.99 \\
MD & 83.26 & \textbf{55.38} & 60.36 & 45.39 \\
AD+MD & 83.04 & 47.22 & 65.03 & 57.44 \\
LB+AD+MD & \textbf{83.63} & 49.19 & \textbf{71.20} & \textbf{59.73} \\ \bottomrule
\end{tabular}
\end{table}

\begin{table}[ht]
\centering
\scriptsize
\caption{Comparisons of different semantic information in GZSSAR accuracies (\%) and harmonic mean.}
\label{sem GZSL}
\vspace{-1mm}
\begin{tabular}{@{}ccccccccccccc@{}}
\toprule
\multirow{4.3}{*}{\textbf{Method}} & \multicolumn{6}{c}{\textbf{NTU-60}} & \multicolumn{6}{c}{\textbf{NTU-120}} \\ \cmidrule(lr){2-7} \cmidrule(lr){8-13}
 & \multicolumn{3}{c}{55($s$)/5($u$)} & \multicolumn{3}{c}{48($s$)/12($u$)} & \multicolumn{3}{c}{110($s$)/10($u$)} & \multicolumn{3}{c}{96($s$)/24($u$)} \\ \cmidrule(lr){2-4} \cmidrule(lr){5-7} \cmidrule(lr){8-10} \cmidrule(lr){11-13}
 & $Acc_s$ & $Acc_u$ & $H$ & $Acc_s$ & $Acc_u$ & $H$ & $Acc_s$ & $Acc_u$ & $H$ & $Acc_s$ & $Acc_u$ & $H$ \\ \midrule
LB & 69.41 & 57.15 & 62.69 & 53.25 & 34.43 & 41.82 & 56.45 & 58.38 & \textbf{57.40} & 58.96 & 35.71 & 44.48 \\
AD & 67.34 & 60.69 & 63.84 & 59.42 & 37.52 & 46.00 & 49.87 & 52.87 & 51.33 & 59.66 & 33.45 & 42.87 \\
MD & 65.04 & 66.74 & 65.88 & 50.69 & \textbf{48.75} & \textbf{49.70} & 58.67 & 52.38 & 55.35 & 58.76 & 32.86 & 42.15 \\
AD+MD & 64.50 & \textbf{72.20} & 68.13 & 57.00 & 39.54 & 46.69 & 45.86 & 62.74 & 52.99 & 50.91 & 51.90 & 51.40 \\
LB+AD+MD & 71.73 & 66.15 & \textbf{68.83} & 58.80 & 40.00 & 47.61 & 46.84 & \textbf{68.30} & 55.57 & 56.84 & \textbf{48.61} & \textbf{52.40} \\ \bottomrule
\end{tabular}
\end{table}

\vspace{-1.2cm}
\subsubsection{Comparisons of Different Semantic Information:}
\label{semantic experiment}
In Section~\ref{semantic information}, three different types of semantic information ($i.e.$, class label (LB), action description (AD) and motion description (MD)) are evaluated.
To verify the advantages of the fusion strategy ($i.e.$, LB+AD+MD), the ablation studies with ZSSAR and GZSSAR settings are conducted.
As shown in Table~\ref{sem ZSL} and Table~\ref{sem GZSL}, generally, the fusion strategy performs superior in both ZSSAR and GZSSAR.
However, under the 48($s$)/12($u$) split of NTU-60 dataset, MD performs even better than fusion.
Because under this split which is proposed in the previous work~\cite{gupta2021syntactically}, unseen classes have less semantical relations with seen classes when using LB or AD as semantic information.
However, MD is not affected because it has more shared words.
And under the 110($s$)/10($u$) split of NTU-120 dataset, LB performs better than fusion in $H$ of GZSSAR.
In this condition, compared to the fusion strategy, $H$ benefits more from $Acc_s$ than from $Acc_u$ when using LB, due to the instability of the seen-unseen classification gate.
Overall, the fusion strategy is the optimal choice.

\vspace{-0.35cm}
\section{Conclusion}
\vspace{-0.15cm}
In this paper, the action descriptions and the motion descriptions are explored for the NTU-60 and NTU-120 datasets. 
The multi-semantic fusion (MSF) model is proposed to integrate multiple semantic information together and accomplish the alignment between skeleton and text features.
On two large-scale datasets ($i.e.$, NTU-60 and NTU-120), the experimental results show that the MSF method outperforms other state-of-the-art methods in both ZSSAR and GZSSAR.
\vspace{-0.3cm}

\nocite{*}
\bibliography{ref}
\bibliographystyle{splncs04.bst}

\end{document}